\documentclass[11pt,a4paper]{article}
\usepackage[hyperref]{emnlp2020}
\usepackage{times}
\usepackage{amsmath,graphicx}
\usepackage{latexsym}

\usepackage{soul}
\usepackage{color}
\usepackage{mathrsfs}
\usepackage{algpseudocode}
\usepackage{multirow}
\usepackage{subfigure}
\usepackage{microtype}
\usepackage{pifont}
\usepackage{bm}
\usepackage{multirow}
\usepackage{graphicx}
\usepackage{booktabs}
\usepackage{tikz}
\usepackage{amssymb}
\usepackage{textcomp}
\usepackage{colortbl}
\usepackage{bibentry}
\usepackage{ulem}
\usepackage{float}
\usepackage{arydshln}
\usepackage{threeparttable}

\colorlet{soulred}{red!50}
\DeclareRobustCommand{\hlred}[1]{{\sethlcolor{soulred}\hl{#1}}}

\colorlet{soulbleu}{cyan!20}
\DeclareRobustCommand{\hlblue}[1]{{\sethlcolor{soulbleu}\hl{#1}}}

\colorlet{soulgreen}{green!20}
\DeclareRobustCommand{\hlgreen}[1]{{\sethlcolor{soulgreen}\hl{#1}}}

\colorlet{soulyellow}{yellow!40}

\colorlet{soulorange}{orange!30}
\DeclareRobustCommand{\hlorange}[1]{{\sethlcolor{soulorange}\hl{#1}}}

\colorlet{soulpurple}{blue!50}

\usepackage{microtype}

\aclfinalcopy 

\title{Multi-hop Question Generation with Graph Convolutional Network}

\author{Dan Su, Yan Xu, Wenliang Dai, Ziwei Ji, Tiezheng Yu, Pascale Fung\\
Center for Artificial Intelligence Research (CAiRE)\\
  Department of Electronic and Computer Engineering\\
  The Hong Kong University of Science and Technology, Clear Water Bay, Hong Kong\\
  \small\texttt{\{dsu, yxucb, wdaiai, zjiad, tyuah\}@connect.ust.hk},\\ \small\texttt{pascale@ece.ust.hk}}

\date{}

\begin{document}
\maketitle
\begin{abstract}
Multi-hop Question Generation (QG) aims to generate answer-related questions by \textit{aggregating} and \textit{reasoning} over multiple scattered evidence from different paragraphs. It is a more challenging yet under-explored task compared to conventional single-hop QG, where the questions are generated from the sentence containing the answer or nearby sentences in the same paragraph without complex reasoning. To address the additional challenges in multi-hop QG, we propose Multi-Hop Encoding Fusion Network for Question Generation (MulQG), which does context encoding in multiple hops with Graph Convolutional Network and encoding fusion via an Encoder Reasoning Gate. To the best of our knowledge, we are the first to tackle the challenge of multi-hop reasoning over paragraphs without any sentence-level information. Empirical results on HotpotQA dataset demonstrate the effectiveness of our method, in comparison with baselines on automatic evaluation metrics. Moreover, from the human evaluation, our proposed model is able to generate fluent questions with high completeness and outperforms the strongest baseline by 20.8\% in the multi-hop evaluation. The code is publicly available at \href{https://github.com/HLTCHKUST/MulQG}{https://github.com/HLTCHKUST/MulQG}.


\end{abstract}




\section{Introduction}
Question Generation (QG) is a task to automatically generate a question from a given context and, optionally, an answer. Recently, we have observed an increasing interest in text-based QG~\cite{du2017learning,zhao2018paragraph,scialom2019self,nema2019let,zhang2019addressing}. 


\begin{table}[!t]
    \centering
    \setlength{\tabcolsep}{3.3pt} 
    \begin{tabular}{p{7.5cm}}
    \hline\hline
    \textbf{Paragraph A:} \hlblue{Marine Tactical Air Command Squadron 28 (\textit{\small Location T})} is a United States Marine Corps
aviation command and control unit based at \hlorange{Marine Corps Air Station Cherry Point (\textit{\small Location C})} ... \\
    \textbf{Paragraph B:} \hlorange{Marine Corps Air Station Cherry Point (\textit{\small Location C})} ... is a United States Marine Corps
airfield located in \hlgreen{Havelock, North Carolina (\textit{\small Location H})}, USA ... \\ 
    \textbf{Answer:}  \hlgreen{Havelock, North Carolina (\textit{\small Location H})} \\ \hline
    \textbf{Question:} What city is the \hlblue{Marine Air Control Group 28 (\textit{\small Location T})} located in? \\ 
    \hline\hline
    \end{tabular}
    \caption{An example of multi-hop QG in the HotpotQA~\cite{yang2018hotpotqa} dataset. Given the answer is Location $H$, to ask where is $T$ located, the model needs a bridging evidence to know that $T$ is located in $C$, and $C$ is located in $H$ ($T \rightarrow C \rightarrow H$). This is done by multi-hop reasoning.
    }
    \label{example}
\end{table}



Most of the existing works on text-based QG focus on generating SQuAD-style~\cite{rajpurkar2016squad, puri2020training} questions, which are generated from the sentence containing the answer or nearby sentences in the same paragraph, via single-hop reasoning~\cite{zhou2017neural,zhao2018paragraph}. Little effort has been put in multi-hop QG, which is a more challenging task. Multi-hop QG requires \textit{aggregating} several scattered evidence spans from multiple paragraphs, and \textit{reasoning} over them to generate answer-related, factual-coherent questions. It can serve as an essential component in education systems~\cite{heilman-smith-2010-good, lindberg2013generating,yao2018teaching}, or be applied in intelligent virtual assistant systems~\cite{shum2018eliza,pan2019reinforced}. It can also combine with question answering (QA) models as dual tasks to boost QA systems with reasoning ability~\cite{tang2017question}.


Intuitively, there are two main additional challenges needed to be addressed for multi-hop QG. The first challenge is how to effectively identify scattered pieces of evidence that can connect the reasoning path of the answer and question~\cite{chauhan2020reinforced}. As the example shown in Table~\ref{example}, to generate a question asking about \textit{``Marine Air Control Group 28''} given only the answer \textit{``Havelock, North Carolina''}, we need the bridging evidence like \textit{``Marine Corps Air Station Cherry Point''}.
The second challenge is how to reason over multiple pieces of scattered evidence to generate factual-coherent questions. 


Previous works mainly focus on single-hop QG, which use neural network based approaches with the sequence-to-sequence (Seq2Seq) framework. Different architectures of encoder and decoder have been designed~\cite{nema2019let, zhao2018paragraph} to incorporate the information of answer and context to do single-hop reasoning. To the best of our knowledge, none of the previous works address the two challenges we mentioned above for multi-hop QG task. The only work on multi-hop QG~\cite{chauhan2020reinforced} uses multi-task learning with an auxiliary loss for sentence-level supporting fact prediction, requiring supporting fact sentences in different paragraphs being labeled in the training data. While labeling those supporting facts requires heavy human labor and is time-consuming, their method cannot be applied to general multi-hop QG cases without supporting facts.


\begin{figure*}[!ht]
 \centering
 \includegraphics[width=0.9\linewidth]{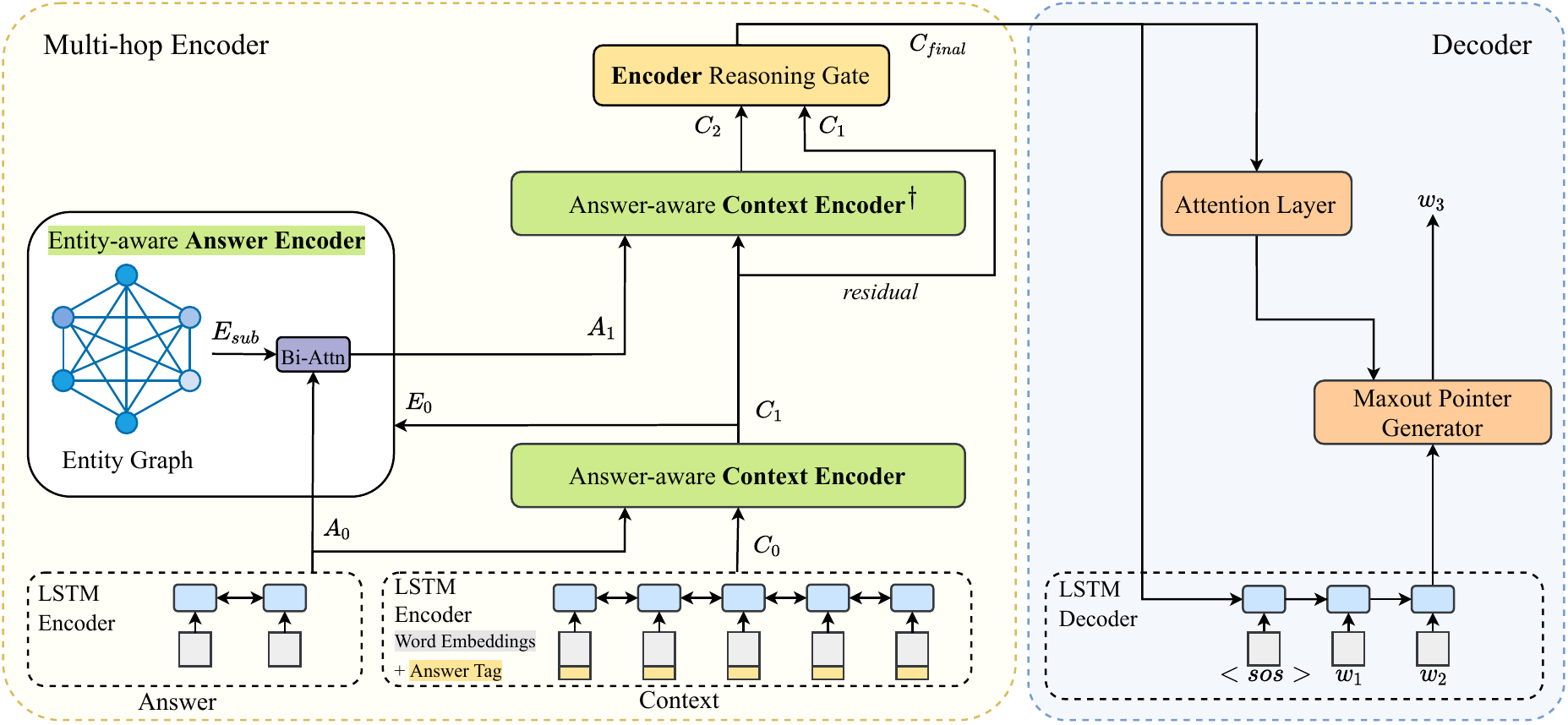}
  \caption{Overview of our MulQG framework. In the encoding stage, we pass the initial context encoding $C_0$ and answer encoding $A_0$ to the \textit{Answer-aware Context Encoder} to obtain the first context encoding $C_1$, then $C_1$ and $A_0$ will be used to update a multi-hop answer encoding $A_1$ via the \textit{GCN-based Entity-aware Answer Encoder}, and we use $A_1$ and $C_1$ back to the \textit{Answer-aware Context Encoder$^\dagger$} to obtain $C_2$. The final context encoding $C_{final}$ are obtained from the \textit{Encoder Reasoning Gate} which operates over $C_1$ and $C_2$, and will be used in the max-out based decoding stage.}
  \label{model}
\end{figure*}

In this paper, we propose a novel architecture named Multi-Hop Encoding Fusion Network for Question Generation (MulQG) to address the aforementioned challenges for multi-hop QG. First of all, it extends the Seq2Seq QG framework from sing-hop to multi-hop for context encoding. Additionally, it leverages a Graph Convolutional Network (GCN) on an answer-aware dynamic entity graph, which is constructed from entity mentions in answer and input paragraphs, to aggregate the potential evidence related to the questions. Moreover, we use different attention mechanisms to imitate the reasoning procedures of human beings in multi-hop generation process, the details are explained in Section~\ref{Methodology}.


We conduct the experiments on the multi-hop QA dataset HotpotQA~\cite{yang2018hotpotqa} with our model and the baselines. The proposed model outperforms the baselines with a significant improvement on automatic evaluation results, such as BLEU~\cite{papineni2002bleu}. The human evaluation results further validate that our proposed model is more likely to generate multi-hop questions with high quality in terms of \textit{Fluency}, \textit{Answerability} and \textit{Completeness} scores. 

Our contributions are summarized as follows:
\begin{itemize}
\setlength{\itemsep}{0pt}
\setlength{\parsep}{0pt}
\setlength{\parskip}{0pt}
\item To the best of our knowledge, we are the first to tackle the challenge of multi-hop reasoning over paragraphs without any sentence-level information in QG tasks. 
\item We propose a new and effective framework for Multi-hop QG, to do context encoding in multiple hops(steps) with Graph Convolutional Network (GCN).
\item We show the effectiveness of our method on both automatic evaluation and human evaluation, and we make the first step to evaluate the model performance in multi-hop aspect.
\end{itemize}

\section{Methodology} \label{Methodology}


The intuition is drawn from human's multi-hop question generation process ~\cite{davey1986effects}. Firstly, given the answer and context, we skim to establish a general understanding of the texts. Then, we find the mentions of entities in or correlated to the answer from the context, and analyse nearby sentences to extract useful evidence. Besides, we may also search for linked information in other paragraphs to gain a further understanding of the entities. Finally, we coherently fuse our knowledge learned from the previous steps and start to generate questions.

To mimic this process, we develop our \textbf{MulQG framework}. The encoding stage is achieved by a novel \textbf{Multi-hop Encoder}. At the decoding stage, we use maxout pointer decoder as proposed in \citet{zhao2018paragraph}. The overview of the framework is shown in Figure~\ref{model}.


\subsection{Multi-hop Encoder}

Our Multi-hop Encoder includes three modules: (1) Answer-aware context encoder (2) GCN-based entity-aware answer encoder (3) Gated encoder reasoning layer.


The context and answer are split into word-level tokens and denoted as $c = \{c_1, c_2, ..., c_n\}$ and $a = \{a_1, a_2, ...,a_m\}$, respectively. Each word is represented by the pre-trained GloVe embedding~\cite{pennington2014glove}. Furthermore, for the words in context, we also append the answer tagging embeddings as described in \citet{zhao2018paragraph}. The context and answer embeddings are fed into two bidirectional LSTM-RNNs separately to obtain their initial contextual representations $C_0 \in R^{d\times n} $ and $A_0 \in R^{d \times m}$, in which $d$ is the hidden state dimension in LSTM.

\begin{figure}[ht]
 \centering
 \includegraphics[width=.9\linewidth]{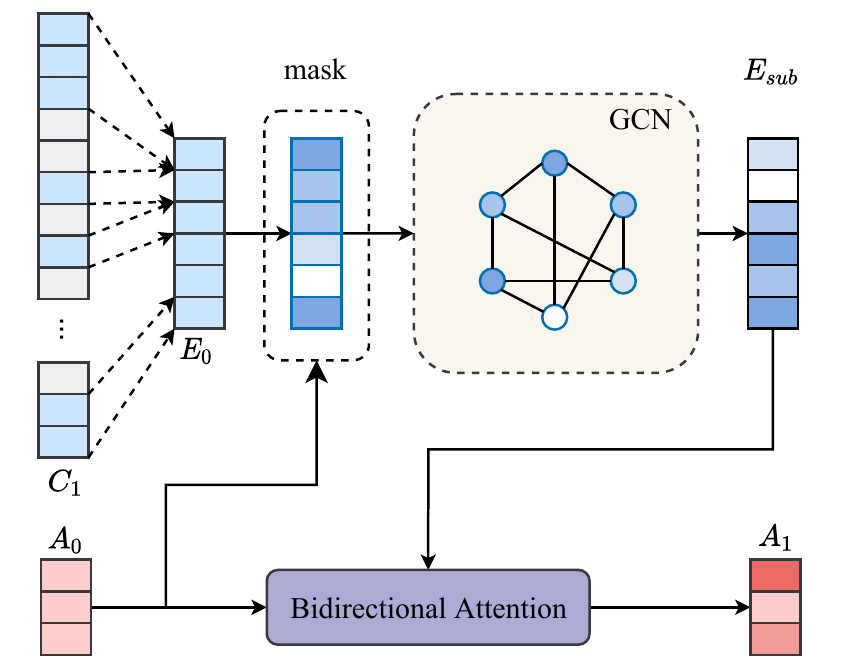}
  \caption{The illustration of GCN-based Entity-aware Answer Encoder.}
  \label{graph}
\end{figure}

\subsubsection{Answer-aware Context Encoder}

Inspired by the co-attention reasoning mechanism in previous machine reading comprehension works~\cite{xiong2016dynamic}, we compute the answer-aware context representation via the following steps:

\begin{align}
    S &= C_0^TA_0 \in R^{n\times m} \label{eq:1} \\
    S' &= \text{softmax}(S)  \in R^{n\times m} \label{eq:2}\\
    S'' &= \text{softmax}(S^T) \in R^{m\times n} \label{eq:3} \\
    A^\prime_0 &= C_0\cdot S' \in R^{d\times m} \label{eq:4} \\
    \tilde{C_1} &= [A_0; A^\prime_0]\cdot S'' \in R^{2d \times n} \label{eq:5}
\end{align}

\begin{equation}
    C_1 = \text{BiLSTM}([\tilde{C_1} ; C_0]) \in R^{d \times n} \label{eq:6}
\end{equation}

Firstly, we compute an alignment matrix $S$ (Eq.\ref{eq:1}), and normalize it column-wise and row-wise to get two attention matrices $S'$ (Eq.\ref{eq:2}) and $S''$ (Eq.\ref{eq:3}). $S'$ represents the relevance of each answer token over the context, and $S''$  represents the relevance of each context token over the answer.
The new answer representation $A^\prime_0$ w.r.t. the context is obtained by Eq.\ref{eq:4}. Next, the answer dependent context representation is calculated by concatenating old and new answer representations and times the attention weight matrix $S''$ (Eq.\ref{eq:5}). 
Finally, to deeply incorporate the interaction between answer and context, we feed the answer dependent representation $\tilde{C_1}$ combined with original $C_0$ into a bi-directional LSTM and obtain the answer-aware context encoding $C_1$ (Eq.\ref{eq:6}).

\subsubsection{GCN-based Entity-aware Answer Encoder}
\label{sec:gcn}
As shown in Figure~\ref{graph}, in order to obtain the multi-hop answer representation, we first compute the entity encoding from the answer-aware context encoding $C_1$, then we apply GCN to propagate multi-hop information on the answer-aware sub-graph. Finally we obtain the updated answer encoding $A_1$ via bi-attention mechanism.

\paragraph{Entity Graph Construction} The entity graph is constructed with the name entities in context as nodes, where we use BERT-based name entity recognition model to recognize name entities from the context. The edges are created for the entity pairs if they are in the same sentence, or appear in the same paragraphs. We also connect the entities from each paragraph title to entities within the same paragraph.

\paragraph{Entity Encoding} With the answer-aware context encoding $C_1$ obtained from Answer-aware Context Encoder, we use a mapping matrix $M$ to calculate the entity encoding. $M$ is a binary matrix where $M_{i,j} = 1$ if the i-th token in the context is within the span of the j-th entity. Each entity's encoding will be calculated via a mean-max pooling applied over it's corresponding context token encoding span. $E_0 = \{e_1, e_2, ...,e_g\} \in R^{2d \times g}$, where $g$ is the number of entities, and $2d$ is the dimension since we directly concatenate the mean-pooling and max-pooling encoding.

\paragraph{Answer-aware GCN}
First we calculate an answer-aware sub-graph, where irrelevant entities are masked out, only those entity nodes related to answer are allowed to disseminate information. Similar to \citet{xiao2019dynamically}, a soft mask $M = [m_1, m_2, ..., m_g]$ is calculated via Eq.~\ref{m}, where each $m_i$ indicate the relatedness of the entity $i$ to the answer, and then apply $M$ on the original graph entities to obtain answer-aware dynamic sub entities graph $E_{sub}$ via Eq.~\ref{esub}.
\begin{align}
\label{m} M &= \sigma(a_0^T\cdot V \cdot E_0) \in R^{1 \times g}\\
\label{esub} E_{sub} &= M\cdot E_0
\end{align} 
where $V$ is a linear projection matrix and $a_0$ is the mean pooling over answer encoding $A_0$, and $\sigma$ is sigmoid function.

Then we calculate the answer-aware sub-graph's attention matrix as described in \citet{velivckovic2017graph} $A_G = \{\alpha_{i,j}\} \in R^{g \times g}$, where $\alpha_{i,j}$ represents the information that will be assigned from entity $i$ to it's neighbor $j$, and obtain the one-layer information propagation over the sub-graph via:
\begin{equation}
\label{e1} E_1 = \text{ReLU}(A_G\cdot E_{sub})
\end{equation}
The computation from Eq.~\ref{e1} can be repeated for multiple times to obtain multi-hop entity representation $E_M$.

\begin{table*}[ht]
\resizebox{\textwidth}{!}{%
\begin{tabular}{c|cccccc|c|l}
\hline
\multirow{2}{*}{\textbf{Model}} & \multicolumn{6}{c|}{\textbf{n-gram}} & \multirow{2}{*}{\textbf{QBLEU4}} & \multirow{2}{*}{\textbf{\begin{tabular}[c]{@{}l@{}}Answer-\\ ability\end{tabular}}} \\ \cline{2-7}
 & \textbf{BLEU-1} & \textbf{BLEU-2} & \textbf{BLEU-3} & \textbf{BLEU-4} & \textbf{ROUGE-L} & \textbf{METEOR} &  &  \\ \hline
 \textbf{\begin{tabular}[c]{@{}c@{}}RefNet$^1$ \\ ~\cite{nema2019let}\end{tabular}} & 29.79 & 19.58 & 14.41 & 11.10 & 30.94 & 18.59 & 51.80 & 70.40 \\ 
 \hdashline
\textbf{\begin{tabular}[c]{@{}c@{}}MP-GSN$*$\\
~\cite{zhao2018paragraph}\end{tabular}} & 34.38 & 23.00 & 17.05 & 13.18 & 31.85 & 19.67 & 48.10 & 64.60 \\
\hline
\textbf{MulQG} & 40.08 & 26.58 & 19.61 & 15.11 & 35.35 & 20.24 & 53.90 & 72.70 \\
\hdashline
\textbf{MulQG + BFS loss} & \textbf{40.15} & \textbf{26.71} & \textbf{19.73} & \textbf{15.20} & \textbf{35.30} & \textbf{20.51} & \textbf{54.00} & \textbf{72.80} \\ \hline
\end{tabular}%
}
\caption{Performance comparison between our MultQG model and state-of-the-art QG models on HotpotQA test set. $^1$The results are obtained with the original implementation of RefNet model. We also follows all the hyper-parameter settings as they are described in the paper.}
\label{results}
\end{table*}

\begin{table*}[ht]
\resizebox{\textwidth}{!}{%
\begin{tabular}{l|cccccc|c|l}
\hline
\multicolumn{1}{c|}{\multirow{2}{*}{\textbf{Setting}}} & \multicolumn{6}{c|}{\textbf{n-gram}} & \multirow{2}{*}{\textbf{QBLEU4}} & \multirow{2}{*}{\textbf{\begin{tabular}[c]{@{}l@{}}Answer-\\ ability\end{tabular}}} \\ \cline{2-7}
 & \textbf{BLEU-1} & \textbf{BLEU-2} & \textbf{BLEU-3} & \textbf{BLEU-4} & \textbf{ROUGE-L} & \textbf{METEOR} &  &  \\ \hline
\textbf{\begin{tabular}[c]{@{}c@{}}MulQG\\ (our model)\end{tabular}} & 40.08 & 26.58 & 19.61 & 15.11 & 35.35 & 20.24 & 53.90 & 72.70 \\ \hline
\textbf{\begin{tabular}[c]{@{}c@{}}MulQG\\  (1-layer GCN)\end{tabular}} & 37.55 & 25.44 & 18.95 & 14.70 & 34.21 & 20.56 & 53.60 & 72.10 \\ \hline
\textbf{w/o GEAEnc} & 36.62 & 24.80 & 18.50 & 14.36 & 33.53 & 20.39 & 52.10 & 70.50 \\ \hline
\textbf{w/o GEAEnc + ACEnc} & 37.85 & 26.19 & 20.15 & 16.21 & 33.35 & 17.86 & 53.40 & 71.90 \\ \hline
\textbf{w/o ERG} & 36.33 & 24.47 & 18.14 & 14.01 & 33.44 & 20.28 & 53.20 & 71.70 \\ \hline
\textbf{w/o GEAEnc + ACEnc + ERG} & 34.01 & 22.95 & 17.09 & 13.26 & 31.90 & 19.90 & 52.40 & 70.70 \\ \hline
\end{tabular}%
}
\caption{Ablation Study of QG performances on HotpotQA test set, with different encoder modules removed. (Here \textbf{GEAEnc}: \textit{\textbf{G}raph-based \textbf{E}ntity-aware \textbf{A}nswer \textbf{Enc}oder},  \textbf{ACEnc}: \textit{\textbf{A}nswer-aware \textbf{C}ontext \textbf{Enc}oder$^\dagger$}, \textbf{ERG}: \textit{\textbf{E}ncoder \textbf{R}easoning \textbf{G}ate)}}
\label{ablation}
\end{table*}



\paragraph{Multi-hop Answer Encoding}
we use bi-attention mechanism \cite{seo2016bidirectional} regarding entities on the sub-graph as memories to update our multi-hop answer encoding $A_1$ via:
\begin{equation}
\setlength\abovedisplayskip{3pt}
\setlength\belowdisplayskip{3pt}
    A_1 = \text{BiAttention}(A_0, E_M)
\end{equation}

\subsubsection{Encoder Reasoning Gate}
We apply a gated feature fusion module on the answer-aware context representations $C_1$ and $C_2$ from previous context encoder hops, to keep and forget information to form the final context representation $C_{final}$ via:
\begin{align}
C_{final}&=g_t\odot C_1\!+\!(1\!-\!g_t) \odot C_2 \\
g_t&=\sigma(w_2^TC_2\!+\!w_1^TC_1\!+\!w_0^TC_0\!+\!b)
\end{align}

\subsection{Maxout Pointer Decoder}
Uni-directional LSTM model is utilized as the decoder of our model. Moreover, we introduce the Maxout Pointer proposed by \citet{zhao2018paragraph} into the decoder for sake of reducing the repetitions in the generation. Pointer Generator enables the decoder to generate the next output token by either computing from the generative probabilistic distribution over the vocabulary or copying from the input sequence. To compute the copy score, the attention over the input sequence which has a vocabulary of $V$ from the current decoder hidden state is leveraged. For the Maxout Pointer Generator, instead of leveraging all the attention score over the input tokens, only the maximal is taken into consideration to avoid the repetitions caused by the input tokens (as it's shown in Eq.~\ref{maxout}, where $a_{t,k}$ annotates the decoder-encoder attention score).
\begin{equation}
    sc^{copy}=
    \begin{cases}
     \mathop{max}\limits_{k,\text{where }x_k=y_t}{a_{t,k}}& \text{, }y_t \in V \\
    -inf& \text{, otherwise}
    \end{cases}
    \label{maxout}
\end{equation}

\subsection{Breadth-First Search Loss}
In addition to the cross-entropy loss, we also introduce Breadth-First Search (BFS) Loss~\cite{xiao2019dynamically} which is a weakly supervised loss to further assist the training procedure. 
Given the answer entities, we conduct the BFS over the adjacent matrices of the entity graph we build to obtain heuristic masks as a weak supervision signal. The BFS loss is calculated via binary cross-entropy loss between the predicted soft masks $M$ in GCN-based Entity-aware Answer Encoder (Section~\ref{sec:gcn}) and the heuristic masks using Eq.~\ref{eq:bfs} to encourage the model to learn the answer-aware dynamic entity graph better.
\begin{equation}
Loss = L_{CrossEntropy} + \lambda L_{BFS} 
\label{eq:bfs}
\end{equation}
where $\lambda$ here is a heuristic number and can be selected using cross-validation.

\section{Experiment}
\subsection{Dataset}
To demonstrate the performance of our model, we conduct the experiments using HotpotQA \cite{yang2018hotpotqa} dataset in an opposite manner. In the QG task, paragraphs and the answers are considered as input, while the corresponding questions are the expected output. HotpotQA is a multi-hop question answering dataset, which contains Wikipedia-based question-answer pairs, with each question requiring multi-hop reasoning across multiple paragraphs to infer the answer. 
There are mainly two types of multi-hop reasoning in the HotpotQA dataset: \textit{bridge} and \textit{comparison}.
Focusing on the multi-hop ability of our model, we filter out all the \textit{yes/no} data samples in the dataset and run our experiments using the remaining corresponding train and test set, which consists of 73k
questions in the training set and 8k in the test set.

\subsection{Baselines}
Since multi-hop QG has been under explored so far, there are very few existing baselines for our comparison. We choose the following two models because of their high relevance with our task and relatively superior performance:

\paragraph{MP-GSN} is the first QG model to demonstrate a large improvement with paragraph-level inputs for single-hop QG proposed by \citet{zhao2018paragraph}. While they conducted their experiments on SQuAD \cite{rajpurkar2016squad}, we use exactly the same experiment settings provided in their configuration file on HotpotQA dataset.

\paragraph{RefNet} is the first work that has reported results on HotpotQA dataset for QG proposed by \citet{nema2019let}. However, their inputs based on the gold supporting sentences, which contains the facts related to the multi-hop question, and no paragraph-level results have been shown. We experiment with the code they released on paragraphs-level, and test their model's performance on both their validation set and test set of HotpotQA dataset. 

We also fine-tuned large pre-trained language models UniLM~\cite{dong2019unified} and BART~\cite{lewis2019bart} on the multi-hop QG task as comparison benchmark, to further show the effectiveness of our method. The details and the results will be covered in Appendix. 

\begin{table*}[!th]
\centering
\begin{tabular}{l|ccc|c}
\hline
\textbf{Model} & \textbf{Fluency} & \textbf{Answerability} & \textbf{Completeness} & \textbf{Multi-hop} \\ \hline
\textbf{Baseline}& 2.26 (0.50)       & 2.08 (0.87)   & 2.30 (0.79)   & 51.5\%   \\
\textbf{Ours}    & 2.46 (0.43)       & 2.49 (0.61)   & 2.83 (0.33)   & 72.3\%   \\
\hline
\textbf{Human}   & 2.57 (0.43)       & 2.67 (0.41)   & 2.86 (0.26)   & 81.2\%   \\
\hline
\end{tabular}
\caption{The results of Human Evaluation. The mean values and the standard deviations of the first three evaluation scores, along with the percentage of questions assessed as multi-hop type are shown above.}
\label{tab:human}
\end{table*} 

\subsection{Implementation Details}

Our word embeddings are initialized by glove.840B.300d \cite{pennington2014glove} and we keep our vocab size as 45000. We use two-layer bi-directional LSTMs for encoder and two-layer uni-directional LSTMs for decoder, and the hidden size is 300 for all the models. We use stochastic gradient descent (SGD) as the optimizer. The initial learning rate is 0.1, and it is reduced during the training stage using a cosine annealing scheduler \cite{loshchilov2016sgdr}. The batch size is 12 and the beam size is 10. We set the dropout probability for LSTM to 0.2 and 0.3 for GCN. The maximum number of epochs is set to 20. We set the maximum number of entities in each context to 80, and we use a two-layer GCN in our GCN-based answer encoder module. After training the model for 10 epochs, we further fine-tune the MulQG model with the help of BFS loss, where the $\lambda$ in Eq.\ref{eq:bfs} is set to 0.5.


\subsection{Automatic Evaluation}
\subsubsection{Metrics} 
We use the metrics in previous work on single-hop QG to evaluate the generation performance of our model, with n-gram similarity metrics BLEU\footnote{https://github.com/Maluuba/nlg-eval}~\cite{papineni2002bleu}, ROUGE-L \cite{lin2004rouge}, and METEOR using the package released in \citet{lavie2009meteor}. We also quantify the QBLEU4~\cite{nema2018towards} and answerability score of our models, which was shown to correlate significantly better with human judgements~\cite{nema-khapra-2018-towards}. 

\subsubsection{Results and Analysis}
Table~\ref{results} shows the performance of various models on the HotpotQA test set. We report the both results of the experiments on our proposed model before and after fine-tuning with auxiliary BFS loss. As it's shown in the table, our MulQG model perform much better than the two baselines methods, with regard to all those measuring metrics, which indicates that the multi-hop procedure can significantly boost the quality of the encoding representations and thus improve the multi-hop question generation performance. Also the BFS loss can further improve the system performance by encouraging learning the answer-aware dynamic entity graph better, which is a key and bottleneck module in the MulQG model. 

\begin{table*}[th]
\begin{minipage}{.32\linewidth}
\centering
\begingroup
\includegraphics[width=\linewidth]{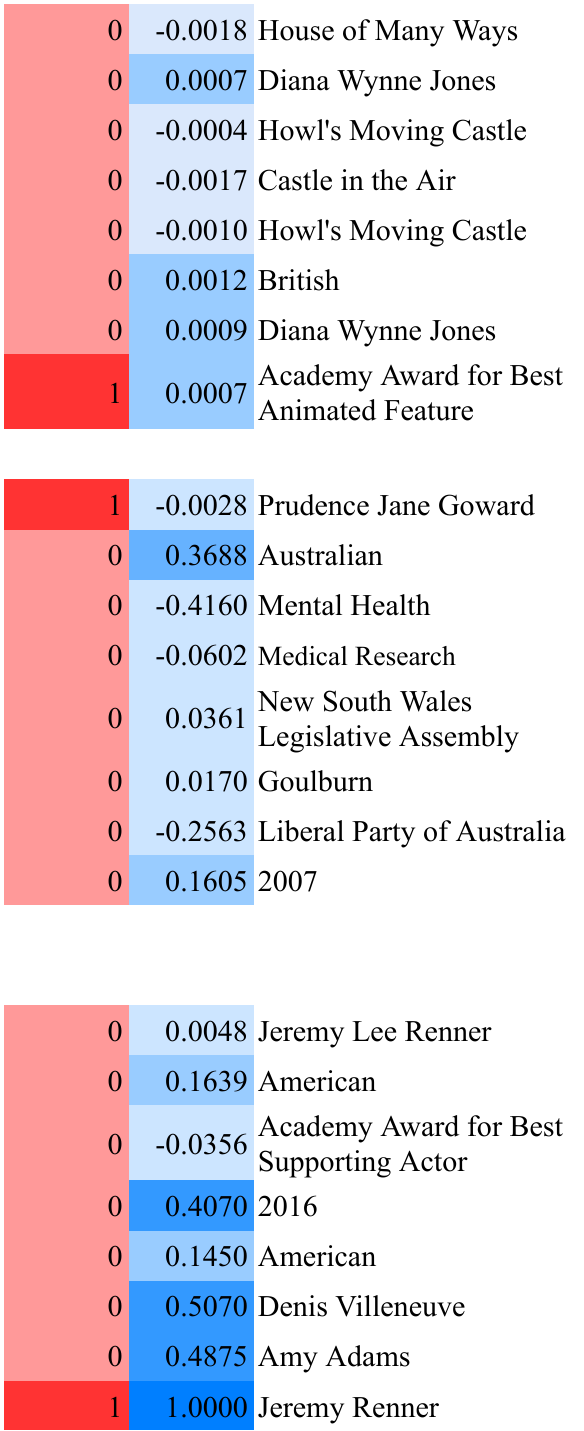}
\endgroup
\end{minipage}%
\begin{minipage}{.67\linewidth}
\centering
\begingroup
\resizebox{1.0\textwidth}{!}{
\begin{tabular}{|lp{13cm}|}
\multicolumn{2}{c}{\textbf{Example \uppercase\expandafter{\romannumeral1}}} \\
\hline
\textbf{Paragraph A:} & \uwave{\textcolor[RGB]{138,190,72}{House of Many Ways} is a young adult fantasy novel written by \textcolor[RGB]{138,190,72}{Diana Wynne Jones}. The story is set in the same world as ``Howl's Moving Castle''} and ``Castle in the Air''.  \\ 
\textbf{Paragraph B:} & Howl's Moving Castle is a fantasy novel by British author \textcolor[RGB]{138,190,72}{Diana Wynne Jones}. ... In 2004 it was adapted as an animated film of the same name, which was \textcolor[RGB]{138,190,72}{nominated} for the academy award for best-animated feature. \\
\hline
\textbf{Answer:} & academy award for best animated feature \\ \hline
\textbf{Baseline:} & house of many ways is a young adult fantasy novel written by diana wynne jones , the story is set in the same world as '' howl 's moving castle \\
\textbf{Ours:} & what award was \textcolor[RGB]{138,190,72}{the author} of the book \textcolor[RGB]{138,190,72}{house of many ways nominated} for ? \\
\textbf{Human: }& house of many ways is a young adult fantasy novel set in the same world as a novel that was adapted as an animated film of the same name and nominated for what ? \\ 
\hline
\multicolumn{2}{c}{\textbf{Example \uppercase\expandafter{\romannumeral2}}} \\
\hline
\textbf{Paragraph A:} & Prudence Jane Goward ( born 2 September 1952 in Adelaide ), an Australian politician, ... \textcolor[RGB]{138,190,72}{she has previously served as the minister for mental health}, minister for medical research, and assistant minister for health between April 2015 and January 2017. ... \textcolor[RGB]{138,190,72}{Goward is a member of the new south wales legislative assembly representing Goulburn} for the liberal party of Australia since 2007.   \\ 
\textbf{Paragraph B:} & \uwave{\textcolor[RGB]{138,190,72}{Goulburn is an electoral district} of the legislative assembly in the Australian state of new south wales. It is represented by Pru Goward of the liberal party of Australia.} \\ \hline
\textbf{Answer:} & jane goward \\ \hline
\textbf{Baseline:} & goulburn is an electoral district of the legislative assembly in the australian state of new south wales , it is represented by pru goward of the liberal party of australia \\
\textbf{Ours:} & which member of \textcolor[RGB]{138,190,72}{the electoral district of goulburn} has \textcolor[RGB]{138,190,72}{previously served as the minister for mental health}? \\
\textbf{Human: }& which australian politician represented electoral district of goulburn \\ \hline
\multicolumn{2}{c}{\textbf{Example \uppercase\expandafter{\romannumeral3}}} \\
\hline
\textbf{Paragraph A:} & Jeremy Lee Renner  (born January 7, 1971) is an \uwave{American actor}. ... He was \textcolor[RGB]{138,190,72}{nominated for the academy award for best supporting actor for his much-praised performance in ``The Town''}.   \\ 
\textbf{Paragraph B:} & \textcolor[RGB]{138,190,72}{Arrival} is a \uwave{2016 American science fiction film directed by Denis Villeneuve} ... It \textcolor[RGB]{138,190,72}{stars} Amy Adams, Jeremy Renner, and Forest Whitaker, ... \\ \hline
\textbf{Answer:} & jeremy renner \\ \hline
\textbf{Baseline:} & which american actor starred in the 2016 american science fiction film directed by denis villeneuve ? \\
\textbf{Ours:} & which star of the movie \textcolor[RGB]{138,190,72}{arrival} was \textcolor[RGB]{138,190,72}{nominated for the academy award for best supporting actor for his performance in `` the town ''}? \\
\textbf{Human: }& name the actor who has acted in the film arrival and who has been nominated for the academy award for best supporting actor for the film `` the town '' ? \\ \hline
\end{tabular}
}
\endgroup
\end{minipage}

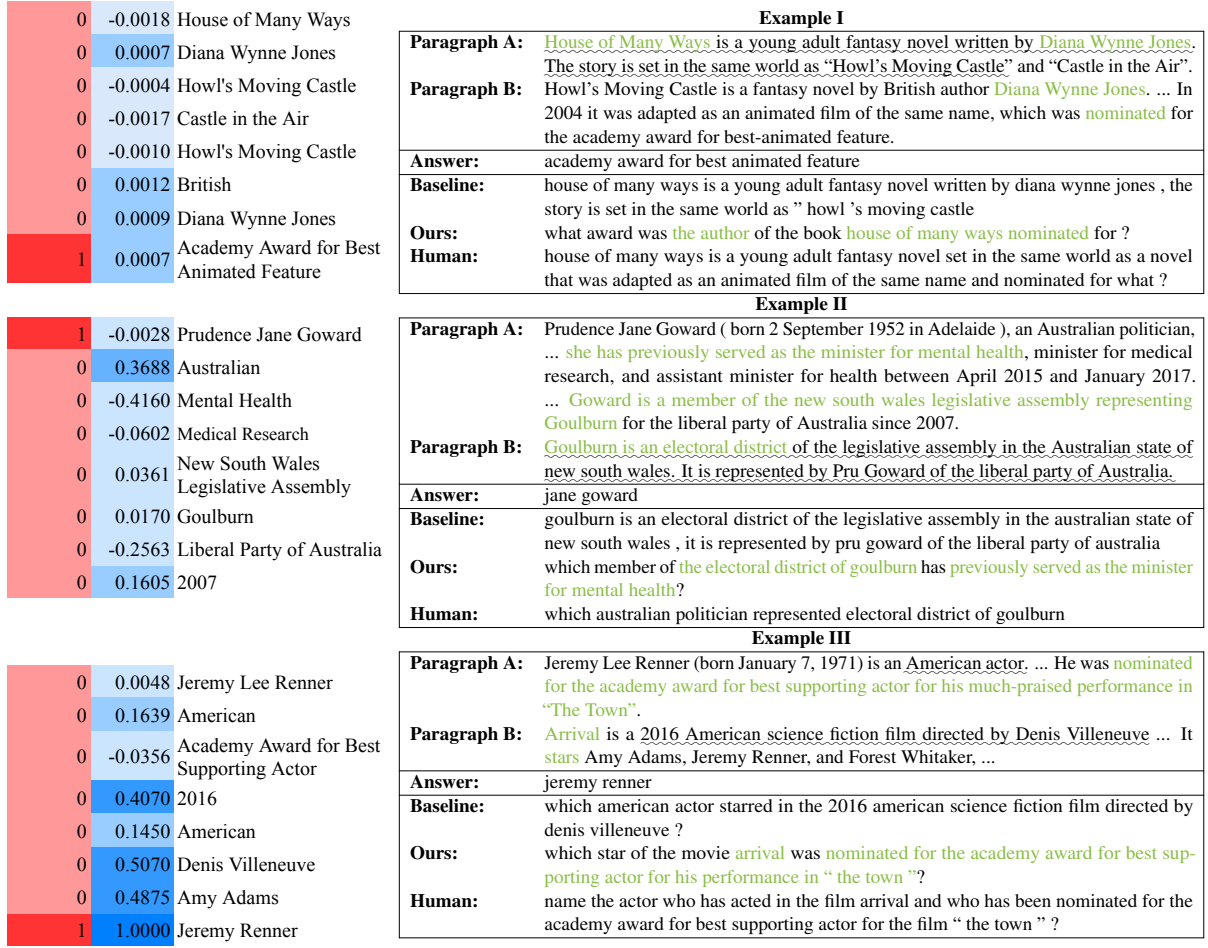
\captionof{figure}{Case study of three examples from the HotpotQA test set. The left part of the figure shows the importance of the entitie nodes, where he left column in \hlred{red} indicates the answer entities and the right colume in \hlblue{blue} displays the importance of the entities of graph reasoning at the starting point by the shade of color. The tables show the generated questions from different models along with the corresponding paragraphs and the answer. Moreover, we highlight the reasoning paths of our proposed model in \textcolor[RGB]{138,190,72}{green} for a more intuitive display. We also use \uwave{wavy lines} to mark out the snippets of the paragraphs that the questions generated by the MP-GSN model derive from.}
\label{case-study}
\end{table*}

\subsubsection{Ablation Study}

To further evaluate and investigate the performance of different components in our model, we perform the ablation study. As we can see from Table~\ref{ablation}, both the \textit{GCN-based entity-aware answer encoder} module and \textit{Gated Context Reasoning} module are important to the model. Each of them provides a relative contribution of 2\%-3\% for overall performance improvement.

\paragraph{w/o \textbf{GEAEnc}:} Without \textit{GCN-based Entity-aware Answer Encoder}, answer-related multi-hop evidence information cannot be identified. Without multi-hop answer encoding being updated, next step's answer-aware context encoding will be affected and thus the performance will drop a lot.

\paragraph{w/o \textbf{GEAEnc} + \textbf{ACEnc}:} The performance continues to decrease but not that much. This matches with our expectation, since without an informative input $A_1$ containing multi-hop information from the \textit{GCN-based Entity-aware Answer Encoder}, the \textit{Answer-aware Context Encoder$^\dagger$} cannot generate an informative $C_2$. Thus remove it won't hurt the performance that much. 

\paragraph{w/o \textbf{ERG}:} When we remove the \textit{Encoder Reasoning Gate}, the performance drops by around 3\% in BLEU-1. This also matches our intuition since without effective feature reasoning and fusion, all the previous encoders cannot generate effective representations. Thus the generation performance will be affected.

\paragraph{w/o \textbf{GEAEnc} + \textbf{ACEnc} + \textbf{ERG}:} Without the three modules, the performance directly drops to single-hop QG system level, which proves the contributions of the whole proposed model.

\paragraph{\textbf{MulQG (1-layer GCN):}} When apply 1-layer GCN and only allow information propagation being limited to each node's neighbor, the answer-related evidences might not be able to be fully obtained, thus the performance are not as good as our 2-layer GCN-based model.

\subsection{Human Evaluation}
Human evaluation is conducted to further analyze the performance of our model (Table~\ref{tab:human}). We compare the generated questions from MP-GSN model, our model and gold ones on four metrics: \textit{\textbf{Fluency}}, \textit{\textbf{Completeness}}, \textit{\textbf{Answerability}} and whether the generated questions are \textit{\textbf{multi-hop question}} or not. Fluency emphasizes the grammar correctness of the question, while Completeness only focuses on the sentence completeness. Answerability mainly indicates the relationship between the answers and the generated questions. For the first three index, the score for each data sample could be chosen from \{1,2,3\} in comparison with the other samples generated from the other two models with the same input, where a higher score indicates a better performance on that matrix, For the multi-hop evaluation, we only carry out binary discrimination. We randomly sample 100 data samples from the test set. Ten annotators are asked in total to evaluate them on the aforementioned four metrics. Each sample is evaluated by three different annotators. 

To present a more convincing analysis, we conduct the t-test on the human evaluation results. All the reported results between our proposed model and the baseline are statistically significant with a p-value$<$0.05. We also calculate the inter-annotator agreement using Fleiss’ Kappa~\cite{fleiss1971measuring} measure and achieve high agreement scores on the proposed model. We observe that our MulQG model largely outperforms the MP-GSN model in terms of Fluency, Answerability and Completeness with more stable quality. Moreover, our model tends to generate more complete question and achieve comparable completeness score with the human annotations. For the multi-hop evaluation, we outperform the strongest baseline by 20.8\% on the multi-hop evaluation. 

\subsection{Case Study}


We present a case study comparing between the strong baseline MP-GSN model, our model and the human annotations. Three cases are presented in Figure~\ref{case-study}. In the first two examples, it’s clearly shown in the examples that the baseline model tends to copy a contiguous and long span of context as the generation, while our proposed model performs better in this aspect. 
we observe that since the supporting fact information is not leveraged in our method, the generated questions from our model may show a different reasoning path with that for the gold question. There could be multiple ways to construct a multi-hop question given the same input. So the generations may be much different from the gold label, although they are still correct questions, which could be indicated from the first two examples. This phenomenon causes a lower score in automatic matrices, such as BLEU and METEOR, but we note that the generated questions still follow the multi-hop scheme and can be answered with the given answers.

In \textit{Example \uppercase\expandafter{\romannumeral3}}, we show the data sample in an easier mode. In this case, while the answer entity is in one paragraph, a similar entity (annotated with orange color) also appears in another paragraph, which gives a strong clue of the reasoning path and makes it easier for the model to attend to both paragraphs. The generations from our model and the human annotation show almost the same reasoning path. However, we observe that the question generated by MP-GSN model still tends to attend to the entities that are closer to the answer entities. Moreover, for the human annotation in \textit{Example \uppercase\expandafter{\romannumeral1}} and \textit{Example \uppercase\expandafter{\romannumeral3}}, the gold questions have a problem with fluency, which is harmful for the QG models, but interestingly, even with training using these labels, our model is still capable of generating relatively fluent outputs. 

\section{Related Work}
\paragraph{Question Generation} Early single-hop QG use rule-based methods to transform sentences to questions~\cite{labutov2015deep, lindberg2013generating}. Recently neural network based approaches adopt the sequence-to-sequence (Seq2Seq) based framework, with different types of encoders and decoders have been designed~\cite{zhou2017neural, nema2019let, zhao2018paragraph}. \citet{zhao2018paragraph} proposes to incorporate paragraph level content by using Gated Self Attention and Maxout pointer networks, while \citet{nema2019let} proposes a model which contains two decoders where the second decoder refines the question generated by the first decoder using reinforcement learning. There are different ways to attend answer information to the context encoding stage. \citet{zhou2017neural} and \citet{liu2019learning} directly concatenate answer tagging with the context embedding, while \citet{nema2019let} also applies bi-attention mechanism proposed by \citet{seo2016bidirectional} for QA to do answer-aware context representation. \citet{chen2019reinforcement} is the most recent work which proposes a reinforcement learning based graph-to-sequence (Graph2Seq) model which use a bidirectional graph encoder on a syntax-based graph for QG, while they still focus on the single-hop QG.

\paragraph{Multi-hop QA} Popular Graph Nueral Network (GNN) frameworks, such as graph convolutional networks~\cite{kipf2016semi},  graph attention network~\cite{velivckovic2017graph}, and graph recurrent network~\cite{song2018graph} have been explored and showed promising results on multi-hop QA task that requiring reasoning. \citet{xiao2019dynamically} proposes a dynamic fused graph network to work on multi-hop QA on the HotpotQA dataset. \citet{de2018question} proposes an entity-GCN method to reason over across multiple documents for multi-hop QA on the WIKIHOP dataset~\cite{welbl2018constructing}.

\section{Conclusion}
Multi-hop QG task is more challenging and worthy of exploration compared to conventional single-hop QG. To address the additional challenges in multi-hop QG, we propose MulQG, which does \textit{multi-hop context encoding} with Graph Convolutional Network and \textit{encoding fusion} via a Gated Reasoning module. To the best of our knowledge, we are the first to tackle the challenge of multi-hop reasoning over paragraphs without any sentence-level information. The model performance on HotpotQA dataset demonstrates its effectiveness on aggregating scattered pieces of evidence across the paragraphs and fusing information effectively to generate multi-hop questions. The strong reasoning ability of the Multi-hop Encoder in the MulQA model can potentially be leveraged in complex generation tasks for the future work.



\normalem
\bibliographystyle{acl_natbib}
\bibliography{emnlp2020}

\begin{thebibliography}{38}
\expandafter\ifx\csname natexlab\endcsname\relax\def\natexlab#1{#1}\fi

\bibitem[{Chauhan et~al.(2020)Chauhan, Ekbal, Bhattacharyya
  et~al.}]{chauhan2020reinforced}
Hardik Chauhan, Asif Ekbal, Pushpak Bhattacharyya, et~al. 2020.
\newblock Reinforced multi-task approach for multi-hop question generation.
\newblock \emph{arXiv preprint arXiv:2004.02143}.

\bibitem[{Chen et~al.(2019)Chen, Wu, and Zaki}]{chen2019reinforcement}
Yu~Chen, Lingfei Wu, and Mohammed~J Zaki. 2019.
\newblock Reinforcement learning based graph-to-sequence model for natural
  question generation.
\newblock \emph{arXiv preprint arXiv:1908.04942}.

\bibitem[{Davey and McBride(1986)}]{davey1986effects}
Beth Davey and Susan McBride. 1986.
\newblock Effects of question-generation training on reading comprehension.
\newblock \emph{Journal of Educational Psychology}, 78(4):256.

\bibitem[{De~Cao et~al.(2018)De~Cao, Aziz, and Titov}]{de2018question}
Nicola De~Cao, Wilker Aziz, and Ivan Titov. 2018.
\newblock Question answering by reasoning across documents with graph
  convolutional networks.
\newblock \emph{arXiv preprint arXiv:1808.09920}.

\bibitem[{Dong et~al.(2019)Dong, Yang, Wang, Wei, Liu, Wang, Gao, Zhou, and
  Hon}]{dong2019unified}
Li~Dong, Nan Yang, Wenhui Wang, Furu Wei, Xiaodong Liu, Yu~Wang, Jianfeng Gao,
  Ming Zhou, and Hsiao-Wuen Hon. 2019.
\newblock Unified language model pre-training for natural language
  understanding and generation.
\newblock In \emph{Advances in Neural Information Processing Systems}, pages
  13063--13075.

\bibitem[{Du et~al.(2017)Du, Shao, and Cardie}]{du2017learning}
Xinya Du, Junru Shao, and Claire Cardie. 2017.
\newblock Learning to ask: Neural question generation for reading
  comprehension.
\newblock In \emph{Proceedings of the 55th Annual Meeting of the Association
  for Computational Linguistics (Volume 1: Long Papers)}, pages 1342--1352.

\bibitem[{Fleiss(1971)}]{fleiss1971measuring}
Joseph~L Fleiss. 1971.
\newblock Measuring nominal scale agreement among many raters.
\newblock \emph{Psychological bulletin}, 76(5):378.

\bibitem[{Heilman and Smith(2010)}]{heilman-smith-2010-good}
Michael Heilman and Noah~A. Smith. 2010.
\newblock \href {https://www.aclweb.org/anthology/N10-1086} {Good question!
  statistical ranking for question generation}.
\newblock In \emph{Human Language Technologies: The 2010 Annual Conference of
  the North {A}merican Chapter of the Association for Computational
  Linguistics}, pages 609--617, Los Angeles, California. Association for
  Computational Linguistics.

\bibitem[{Kipf and Welling(2016)}]{kipf2016semi}
Thomas~N Kipf and Max Welling. 2016.
\newblock Semi-supervised classification with graph convolutional networks.
\newblock \emph{arXiv preprint arXiv:1609.02907}.

\bibitem[{Labutov et~al.(2015)Labutov, Basu, and Vanderwende}]{labutov2015deep}
Igor Labutov, Sumit Basu, and Lucy Vanderwende. 2015.
\newblock Deep questions without deep understanding.
\newblock In \emph{Proceedings of the 53rd Annual Meeting of the Association
  for Computational Linguistics and the 7th International Joint Conference on
  Natural Language Processing (Volume 1: Long Papers)}, pages 889--898.

\bibitem[{Lavie and Denkowski(2009)}]{lavie2009meteor}
Alon Lavie and Michael~J Denkowski. 2009.
\newblock The meteor metric for automatic evaluation of machine translation.
\newblock \emph{Machine translation}, 23(2-3):105--115.

\bibitem[{Lewis et~al.(2019)Lewis, Liu, Goyal, Ghazvininejad, Mohamed, Levy,
  Stoyanov, and Zettlemoyer}]{lewis2019bart}
Mike Lewis, Yinhan Liu, Naman Goyal, Marjan Ghazvininejad, Abdelrahman Mohamed,
  Omer Levy, Ves Stoyanov, and Luke Zettlemoyer. 2019.
\newblock Bart: Denoising sequence-to-sequence pre-training for natural
  language generation, translation, and comprehension.
\newblock \emph{arXiv preprint arXiv:1910.13461}.

\bibitem[{LIN(2004)}]{lin2004rouge}
C-Y LIN. 2004.
\newblock Rouge: A package for automatic evaluation of summaries.
\newblock In \emph{Proc. of Workshop on Text Summarization Branches Out, Post
  Conference Workshop of ACL 2004}.

\bibitem[{Lindberg et~al.(2013)Lindberg, Popowich, Nesbit, and
  Winne}]{lindberg2013generating}
David Lindberg, Fred Popowich, John Nesbit, and Phil Winne. 2013.
\newblock Generating natural language questions to support learning on-line.
\newblock In \emph{Proceedings of the 14th European Workshop on Natural
  Language Generation}, pages 105--114.

\bibitem[{Liu et~al.(2019)Liu, Zhao, Niu, Lai, He, Wei, and
  Xu}]{liu2019learning}
Bang Liu, Mingjun Zhao, Di~Niu, Kunfeng Lai, Yancheng He, Haojie Wei, and
  Yu~Xu. 2019.
\newblock Learning to generate questions by learningwhat not to generate.
\newblock In \emph{The World Wide Web Conference}, pages 1106--1118. ACM.

\bibitem[{Loshchilov and Hutter(2016)}]{loshchilov2016sgdr}
Ilya Loshchilov and Frank Hutter. 2016.
\newblock Sgdr: Stochastic gradient descent with warm restarts.
\newblock \emph{arXiv preprint arXiv:1608.03983}.

\bibitem[{Nema and Khapra(2018{\natexlab{a}})}]{nema2018towards}
Preksha Nema and Mitesh~M Khapra. 2018{\natexlab{a}}.
\newblock Towards a better metric for evaluating question generation systems.
\newblock In \emph{Proceedings of the 2018 Conference on Empirical Methods in
  Natural Language Processing}, pages 3950--3959.

\bibitem[{Nema and Khapra(2018{\natexlab{b}})}]{nema-khapra-2018-towards}
Preksha Nema and Mitesh~M. Khapra. 2018{\natexlab{b}}.
\newblock \href {https://doi.org/10.18653/v1/D18-1429} {Towards a better metric
  for evaluating question generation systems}.
\newblock In \emph{Proceedings of the 2018 Conference on Empirical Methods in
  Natural Language Processing}, pages 3950--3959, Brussels, Belgium.
  Association for Computational Linguistics.

\bibitem[{Nema et~al.(2019)Nema, Mohankumar, Khapra, Srinivasan, and
  Ravindran}]{nema2019let}
Preksha Nema, Akash~Kumar Mohankumar, Mitesh~M Khapra, Balaji~Vasan Srinivasan,
  and Balaraman Ravindran. 2019.
\newblock Let’s ask again: Refine network for automatic question generation.
\newblock In \emph{Proceedings of the 2019 Conference on Empirical Methods in
  Natural Language Processing and the 9th International Joint Conference on
  Natural Language Processing (EMNLP-IJCNLP)}, pages 3305--3314.

\bibitem[{Pan et~al.(2019)Pan, Li, Yao, Cai, and Sun}]{pan2019reinforced}
Boyuan Pan, Hao Li, Ziyu Yao, Deng Cai, and Huan Sun. 2019.
\newblock Reinforced dynamic reasoning for conversational question generation.
\newblock \emph{arXiv preprint arXiv:1907.12667}.

\bibitem[{Papineni et~al.(2002)Papineni, Roukos, Ward, and
  Zhu}]{papineni2002bleu}
Kishore Papineni, Salim Roukos, Todd Ward, and Wei-Jing Zhu. 2002.
\newblock Bleu: a method for automatic evaluation of machine translation.
\newblock In \emph{Proceedings of the 40th annual meeting on association for
  computational linguistics}, pages 311--318. Association for Computational
  Linguistics.

\bibitem[{Pennington et~al.(2014)Pennington, Socher, and
  Manning}]{pennington2014glove}
Jeffrey Pennington, Richard Socher, and Christopher~D Manning. 2014.
\newblock Glove: Global vectors for word representation.
\newblock In \emph{Proceedings of the 2014 conference on empirical methods in
  natural language processing (EMNLP)}, pages 1532--1543.

\bibitem[{Puri et~al.(2020)Puri, Spring, Patwary, Shoeybi, and
  Catanzaro}]{puri2020training}
Raul Puri, Ryan Spring, Mostofa Patwary, Mohammad Shoeybi, and Bryan Catanzaro.
  2020.
\newblock Training question answering models from synthetic data.
\newblock \emph{arXiv preprint arXiv:2002.09599}.

\bibitem[{Rajpurkar et~al.(2016)Rajpurkar, Zhang, Lopyrev, and
  Liang}]{rajpurkar2016squad}
Pranav Rajpurkar, Jian Zhang, Konstantin Lopyrev, and Percy Liang. 2016.
\newblock Squad: 100,000+ questions for machine comprehension of text.
\newblock In \emph{Proceedings of the 2016 Conference on Empirical Methods in
  Natural Language Processing}, pages 2383--2392.

\bibitem[{Scialom et~al.(2019)Scialom, Piwowarski, and
  Staiano}]{scialom2019self}
Thomas Scialom, Benjamin Piwowarski, and Jacopo Staiano. 2019.
\newblock Self-attention architectures for answer-agnostic neural question
  generation.
\newblock In \emph{Proceedings of the 57th Annual Meeting of the Association
  for Computational Linguistics}, pages 6027--6032.

\bibitem[{Seo et~al.(2016)Seo, Kembhavi, Farhadi, and
  Hajishirzi}]{seo2016bidirectional}
Minjoon Seo, Aniruddha Kembhavi, Ali Farhadi, and Hannaneh Hajishirzi. 2016.
\newblock Bidirectional attention flow for machine comprehension.
\newblock \emph{arXiv preprint arXiv:1611.01603}.

\bibitem[{Shum et~al.(2018)Shum, He, and Li}]{shum2018eliza}
Heung-Yeung Shum, Xiao-dong He, and Di~Li. 2018.
\newblock From eliza to xiaoice: challenges and opportunities with social
  chatbots.
\newblock \emph{Frontiers of Information Technology \& Electronic Engineering},
  19(1):10--26.

\bibitem[{Song et~al.(2018)Song, Zhang, Wang, and Gildea}]{song2018graph}
Linfeng Song, Yue Zhang, Zhiguo Wang, and Daniel Gildea. 2018.
\newblock A graph-to-sequence model for amr-to-text generation.
\newblock \emph{arXiv preprint arXiv:1805.02473}.

\bibitem[{Tang et~al.(2017)Tang, Duan, Qin, Yan, and Zhou}]{tang2017question}
Duyu Tang, Nan Duan, Tao Qin, Zhao Yan, and Ming Zhou. 2017.
\newblock Question answering and question generation as dual tasks.
\newblock \emph{arXiv preprint arXiv:1706.02027}.

\bibitem[{Veli{\v{c}}kovi{\'c} et~al.(2017)Veli{\v{c}}kovi{\'c}, Cucurull,
  Casanova, Romero, Lio, and Bengio}]{velivckovic2017graph}
Petar Veli{\v{c}}kovi{\'c}, Guillem Cucurull, Arantxa Casanova, Adriana Romero,
  Pietro Lio, and Yoshua Bengio. 2017.
\newblock Graph attention networks.
\newblock \emph{arXiv preprint arXiv:1710.10903}.

\bibitem[{Welbl et~al.(2018)Welbl, Stenetorp, and
  Riedel}]{welbl2018constructing}
Johannes Welbl, Pontus Stenetorp, and Sebastian Riedel. 2018.
\newblock Constructing datasets for multi-hop reading comprehension across
  documents.
\newblock \emph{Transactions of the Association for Computational Linguistics},
  6:287--302.

\bibitem[{Xiao et~al.(2019)Xiao, Qu, Qiu, Zhou, Li, Zhang, and
  Yu}]{xiao2019dynamically}
Yunxuan Xiao, Yanru Qu, Lin Qiu, Hao Zhou, Lei Li, Weinan Zhang, and Yong Yu.
  2019.
\newblock Dynamically fused graph network for multi-hop reasoning.
\newblock \emph{arXiv preprint arXiv:1905.06933}.

\bibitem[{Xiong et~al.(2016)Xiong, Zhong, and Socher}]{xiong2016dynamic}
Caiming Xiong, Victor Zhong, and Richard Socher. 2016.
\newblock Dynamic coattention networks for question answering.
\newblock \emph{arXiv preprint arXiv:1611.01604}.

\bibitem[{Yang et~al.(2018)Yang, Qi, Zhang, Bengio, Cohen, Salakhutdinov, and
  Manning}]{yang2018hotpotqa}
Zhilin Yang, Peng Qi, Saizheng Zhang, Yoshua Bengio, William Cohen, Ruslan
  Salakhutdinov, and Christopher~D Manning. 2018.
\newblock Hotpotqa: A dataset for diverse, explainable multi-hop question
  answering.
\newblock In \emph{Proceedings of the 2018 Conference on Empirical Methods in
  Natural Language Processing}, pages 2369--2380.

\bibitem[{Yao et~al.(2018)Yao, Zhang, Luo, Tao, and Wu}]{yao2018teaching}
Kaichun Yao, Libo Zhang, Tiejian Luo, Lili Tao, and YanJun Wu. 2018.
\newblock Teaching machines to ask questions.
\newblock In \emph{Proceedings of the 27th International Joint Conference on
  Artificial Intelligence}, pages 4546--4552. AAAI Press.

\bibitem[{Zhang and Bansal(2019)}]{zhang2019addressing}
Shiyue Zhang and Mohit Bansal. 2019.
\newblock Addressing semantic drift in question generation for semi-supervised
  question answering.
\newblock In \emph{Proceedings of the 2019 Conference on Empirical Methods in
  Natural Language Processing and the 9th International Joint Conference on
  Natural Language Processing (EMNLP-IJCNLP)}, pages 2495--2509.

\bibitem[{Zhao et~al.(2018)Zhao, Ni, Ding, and Ke}]{zhao2018paragraph}
Yao Zhao, Xiaochuan Ni, Yuanyuan Ding, and Qifa Ke. 2018.
\newblock Paragraph-level neural question generation with maxout pointer and
  gated self-attention networks.
\newblock In \emph{Proceedings of the 2018 Conference on Empirical Methods in
  Natural Language Processing}, pages 3901--3910.

\bibitem[{Zhou et~al.(2017)Zhou, Yang, Wei, Tan, Bao, and
  Zhou}]{zhou2017neural}
Qingyu Zhou, Nan Yang, Furu Wei, Chuanqi Tan, Hangbo Bao, and Ming Zhou. 2017.
\newblock Neural question generation from text: A preliminary study.
\newblock In \emph{National CCF Conference on Natural Language Processing and
  Chinese Computing}, pages 662--671. Springer.

\end{thebibliography}

\clearpage
\appendix

\setcounter{table}{0} 
\setcounter{figure}{0}
\renewcommand{\thetable}{\Alph{section}\arabic{table}}
\renewcommand\thefigure{\Alph{section}\arabic{figure}} 

\section{Appendix}
\label{sec:appendix}

\begin{table*}[ht]
\resizebox{\textwidth}{!}
{%
\begin{tabular}{c|cccccc}
\hline
 \textbf{Model} & \textbf{BLEU-1} & \textbf{BLEU-2} & \textbf{BLEU-3} & \textbf{BLEU-4} & \textbf{ROUGE-L} & \textbf{METEOR}\\ \hline
 \textbf{\begin{tabular}[c]{@{}c@{}}Finetune-UniLM(l48\_p0\_b1) \\ \end{tabular}} & 42.37 & 29.95 & 22.61 & 17.61 & 40.34 & 25.48 \\  \hdashline

 \textbf{\begin{tabular}[c]{@{}c@{}} Finetune-BART(test.hypo.l32\_p0\_b5) \\ \end{tabular}} &  41.41 & 30.90 & 24.39 & 19.75 & 36.13 & 25.20 \\   
\hline
\textbf{MulQG} & 40.08 & 26.58 & 19.61 & 15.11 & 35.35 & 20.24  \\
\hdashline
\textbf{MulQG + BFS loss} & 40.15 & 26.71 & 19.73 & 15.20 & 35.30 & 20.51  \\ \hline
\end{tabular}%
}
\caption{Performance comparison between our MultQG model and fine-tuning state-of-the-art large pre-trained models on HotpotQA test set.}
\label{results-large_pretrained}
\end{table*}

\subsection{Detailed Experiment Settings}
We run our experiments on 1 GeForce® GTX 1080 Ti GPU, with batch size to 12. The average runtime for our model is around 7500s for one epoch. The total numbers of parameters for our model is : 84250510, while we freeze the word embedding parameters, so our total number of parameters need to be optimized is 57250510. We run the baselines also on the same computing environment, using the configuration file they provided. For the Maxout Pointer baseline, we use a batch size of 16 to fit with our GPU memory.

\subsection{Comparison with fine-tuning large pre-trained language models}

In order to further show the effectiveness of our method, we further fine-tuned UniLM~\cite{dong2019unified} and BART ~\cite{lewis2019bart} on the multi-hop QG task. UniLM and BART has obtained state-of-the-art performance on the summarization tasks and also on question generation task on SQuAD ~\cite{rajpurkar2016squad} dataset.

As we can see from Table~\ref{results-large_pretrained}, the performance of our model is on par with fine-tuning the large-pretrained models on the multihop QG tasks. While our model is much more light-weight and can provide explicit reasoning interpretability.

\end{document}